%% file: acl_latex.tex
\pdfoutput=1

\documentclass[11pt]{article}

\usepackage{acl}

\usepackage{times}
\usepackage{latexsym}

\usepackage[T1]{fontenc}

\usepackage[utf8]{inputenc}

\usepackage{microtype}

\usepackage{amsmath}
\usepackage{amssymb}
\usepackage{url}
\usepackage{booktabs} 
\usepackage{graphicx}
\usepackage{epstopdf}
\usepackage{subfigure}
\usepackage{verbatim}
\usepackage{algpseudocode}
\usepackage{pifont}
\usepackage{bm}
\usepackage{array}
\usepackage{multirow}
\usepackage{subfigure}
\usepackage{makecell}
\usepackage{xcolor}
\usepackage[most]{tcolorbox}

\usepackage[font={small}]{caption}

\title{KETOD: Knowledge-Enriched Task-Oriented Dialogue}

\author{\textbf{Zhiyu Chen}\textsuperscript{1}\thanks{\hspace{5pt}Work done as a research intern at Meta.}, \textbf{Bing Liu}\textsuperscript{2}, \textbf{Seungwhan Moon}\textsuperscript{2}, \textbf{Chinnadhurai Sankar}\textsuperscript{2}, \\ \textbf{Paul Crook}\textsuperscript{2} and \textbf{William Yang Wang}\textsuperscript{1} \\
  \textsuperscript{1}University of California, Santa Barbara \\
  \textsuperscript{2}Meta \\
  {\tt \{zhiyuchen,william\}@cs.ucsb.edu}, \\ {\tt \{bingl,shanemoon,chinnadhurai,pacrook\}@fb.com} \\}

\begin{document}
\maketitle

\input{00-abstract.tex}
\input{01-introduction.tex}
\input{02-related.tex}
\input{03-dataset.tex}
\input{04-model.tex}
\input{05-experiments.tex}

\input{06-conclusion.tex}
\input{07-ethics.tex}

\bibliography{custom}
\bibliographystyle{acl_natbib}

\input{08-appendix.tex}




\end{document}

%% file: 00-abstract.tex
\begin{abstract}
Existing studies in dialogue system research mostly treat task-oriented dialogue and chit-chat as separate domains. Towards building a human-like assistant that can converse naturally and seamlessly with users, it is important to build a dialogue system that conducts both types of conversations effectively. In this work, we investigate how task-oriented dialogue and knowledge-grounded chit-chat can be effectively integrated into a single model.
To this end, we create a new dataset, KETOD (\underline{K}nowledge-\underline{E}nriched \underline{T}ask-\underline{O}riented \underline{D}ialogue), where we naturally enrich task-oriented dialogues with chit-chat based on relevant entity knowledge. We also propose two new models, SimpleToDPlus and Combiner, for the proposed task. Experimental results on both automatic and human evaluations show that the proposed methods can significantly improve the performance in knowledge-enriched response generation while maintaining a competitive task-oriented dialog performance. We believe our new dataset will be a valuable resource for future studies. Our dataset and code are publicly available\footnote{ \url{https://github.com/facebookresearch/ketod})}.
\end{abstract}


%% file: 01-introduction.tex
\section{Introduction}
\begin{figure}[ht]
\centering
\includegraphics[width=0.48\textwidth]{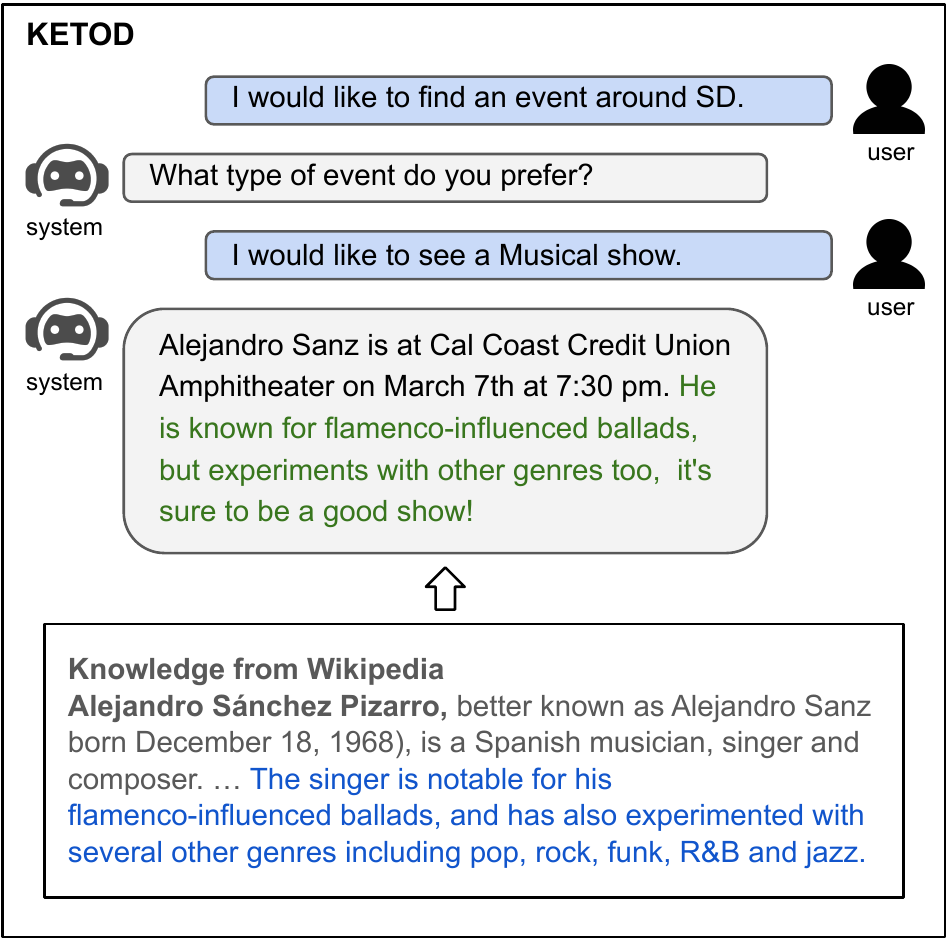}
\caption{An example from the KETOD dataset: the green text is our enriched chit-chat based on the entity knowledge of \textit{Alejandro Sanz} in the original TOD. Such knowledge-grounded chit-chat makes the dialogue more natural and engaging. } 
\label{fig:eg0}
\end{figure}
Dialogue systems have achieved substantial progress~\cite{DBLP:conf/acl/ZhangSGCBGGLD20, DBLP:journals/corr/abs-2005-00796, DBLP:conf/acl/TaoCFW020} due to recent success in language model pre-training~\cite{radford2019language, DBLP:journals/jmlr/RaffelSRLNMZLL20, DBLP:conf/acl/LewisLGGMLSZ20}. One major type of dialogue being studied is task-oriented dialogue (TOD)~\cite{DBLP:conf/icml/WenMBY17, DBLP:conf/emnlp/BudzianowskiWTC18, DBLP:conf/aaai/RastogiZSGK20, DBLP:journals/corr/abs-2005-00796}, where the system aims to collect user intents/goals to complete certain tasks (e.g. restaurant-booking). In most of TOD systems, the system responses are concise and templated, as we only focus on the success of task completion but not providing a natural and engaging conversational experience. The latter is the target of another kind of popularly studied dialogue - knowledge-grounded chit-chat~\cite{DBLP:conf/aaai/GhazvininejadBC18, DBLP:conf/acl/KielaWZDUS18, DBLP:conf/emnlp/TuanCL19, DBLP:conf/iclr/DinanRSFAW19}. 
Knowledge-grounded chit-chat enables dialog systems to access external knowledge so that they can provide more engaging and knowledgeable conversations and in the same time reduce hallucinations~\cite{DBLP:conf/emnlp/0001PCKW21}.

Existing studies mostly focus on one specific type of dialogue, either task-oriented dialogue or knowledge-grounded chit-chat. However, the ultimate goal of Conversational AI is a human-like, unified system capable of conversing with the users naturally and seamlessly among all kinds of dialogues. Current TOD systems can hardly make interesting and engaging conversations only with templated functional responses. Few previous works like ACCENTOR~\cite{DBLP:conf/naacl/SunMCRSLWLCC21} have studied the combination of TOD and chit-chat, but their chit-chat augmentation is largely limited to simple general responses like `you're welcome', `sounds good to me'. In this work, we propose to enrich TOD with knowledge-grounded chit-chat, as one step further towards the ultimate goal of building a human-like, unified system (See Figure~\ref{fig:eg0} for an example). We believe that the proposed knowledge-enriched TOD system can conduct more social, natural, and engaging conversations.

To this end, we propose a new dataset, KETOD (\underline{K}nowledge-\underline{E}nriched \underline{T}ask-\underline{O}riented \underline{D}ialogue). In order to obtain natural and high-quality knowledge-grounded chit-chat, we design the dataset construction framework by augmenting existing TODs and using the relevant entity knowledge to make the chit-chat enrichment. Specifically, for a given TOD, 1) extracting the entities from the dialogue states and actions; 2) retrieving the knowledge associated with the entities from external knowledge sources; 3) asking the human annotators to enrich the system responses with chit-chat using the retrieved knowledge. We demonstrate that the knowledge-enriched dialogues constructed with the proposed framework are consistently preferred by human judges across all axes of engagingness, interestingness, knowledge, and humanness. 

We propose two models, and study the challenges and insights of our new dataset. The first model is an end-to-end language model that jointly learns and generates both the TOD results (dialogue states and actions) and the knowledge-enriched responses. The second model is a pipeline that first generates the TOD results, then uses another response generation model to generate the knowledge-enriched responses. We run comprehensive experiments to demonstrate the improvement over the baselines, and show that our models can generate better knowledge-enriched responses while maintaining competitive performance on the TOD tasks. 
To summarize, we make the following major contributions:

\begin{itemize}
    \item We propose the task of combining TOD and knowledge-grounded chit-chat.
    \item We construct a new large-scale dataset, KETOD, with high-quality, manually annotated dialogue responses enriched with knowledge-grounded chit-chat. We will release the dataset upon acceptance of the paper. 
    \item We propose two models for our dataset, and carry comprehensive experiments to study the challenges and insights. We believe our dataset should be a valuable resource for building a human-like conversational assistant. 

\end{itemize}

%% file: 02-related.tex
\section{Related Work}
\begin{figure*}[ht]
\centering
\includegraphics[width=0.96\textwidth]{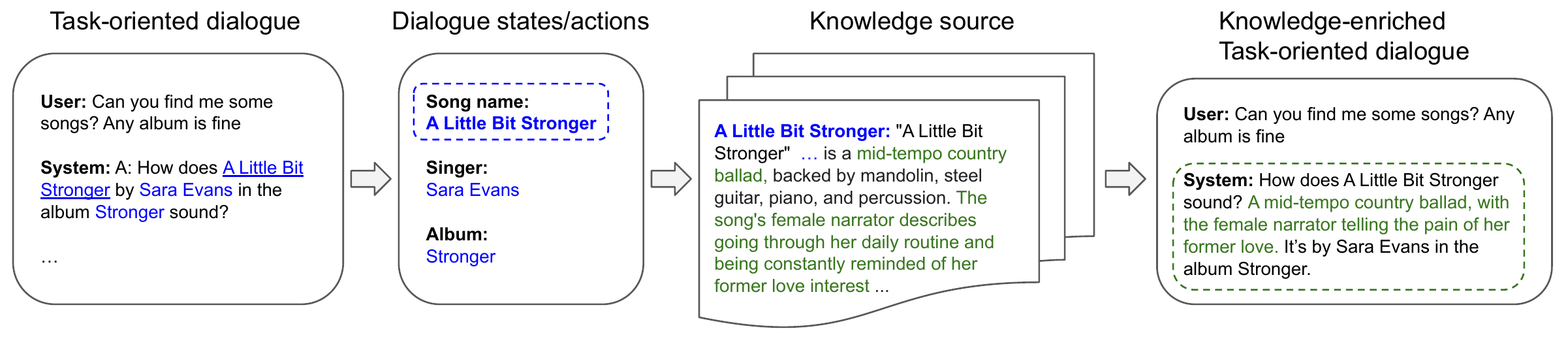}
\caption{The pipeline of dataset construction: for each task-oriented dialogue, we first extract all the entities from the dialogue states and actions. Then we retrieve the knowledge associated with each entity from external knowledge sources (Wikipedia). At last, we ask human annotators to enrich the TOD system responses with chit-chat grounded on the retrieved knowledge. } 
\label{fig:data_pipeline}
\end{figure*}
\noindent \textbf{Task-oriented dialogue.}
Task-oriented dialogue (TOD) has been one of the most popular types of dialogue in the research community. There have been many works on building each component of the TOD system, such as dialogue state tracking, action prediction, and response generation~\cite{DBLP:conf/emnlp/WenGMSVY15, DBLP:conf/eacl/Rojas-BarahonaG17, DBLP:conf/acl/MrksicSWTY17, DBLP:conf/acl/SocherZX18, DBLP:conf/lrec/EricGPSAGKGKH20, DBLP:conf/naacl/LiuTHSH18, DBLP:conf/emnlp/PengLLGCLW17, DBLP:conf/aaai/ZhouLCLCH17}. Later works begin to investigate building end-to-end systems~\cite{DBLP:conf/iclr/BordesBW17, DBLP:conf/naacl/LiuTHSH18, DBLP:journals/corr/abs-1711-10712, DBLP:journals/www/XuPXCZZ20}. Most recent works on TOD also apply such language model pre-training style methods on building end-to-end systems~\cite{DBLP:journals/corr/abs-2005-00796, DBLP:journals/corr/abs-2005-05298, DBLP:journals/corr/abs-2109-14739}, achieving top performances on various datasets. 
Popular datasets in TOD include the DSTC challenge series~\cite{DBLP:journals/dad/WilliamsRH16a}, MultiWOZ~\cite{DBLP:conf/emnlp/BudzianowskiWTC18}, SGD~\cite{DBLP:conf/aaai/RastogiZSGK20}, etc. As the primary goal of TOD is the successful completion of the functional tasks, the system responses are  mostly concise and templated. 

\noindent \textbf{Chit-chat dialogue.} Another type of popular studied dialogue is chit-chat, with the goal of making a natural and engaging conversation. Apart from the `pure' simple chit-chat that mostly covers plain and general responses, more works focus on knowledge groundings to achieve better specificity and engagingness, such as using user profiles~\cite{DBLP:conf/acl/KielaWZDUS18}, social media contexts~\cite{DBLP:conf/naacl/SordoniGABJMNGD15}, or knowledge graphs~\cite{DBLP:conf/emnlp/TuanCL19, DBLP:conf/acl/MoonSKS19}, etc. In this work, our enriched chit-chat is grounded on open-domain knowledge, similar as the Topical-Chat~\cite{DBLP:conf/interspeech/GopalakrishnanH19} and the WOW dataset~\cite{DBLP:conf/iclr/DinanRSFAW19}, where the system converses with the users about certain topics involving entity knowledge in an open-ended setting. In contrast, their datasets specifically focus
on knowledge-grounded chit-chat, while our dataset combines TOD and such chit-chat. 


\noindent \textbf{Combination of task-oriented dialogue and chit-chat.}
ACCENTOR~\cite{DBLP:conf/naacl/SunMCRSLWLCC21} proposes to combine TOD with chit-chat by prepending or appending chit-chat to the TOD system responses. But their chit-chat is mostly general responses like 'sounds good!', 'you're welcome'. 
FusedChat~\cite{DBLP:journals/corr/abs-2109-04137} proposes to insert chit-chat turns into TOD as well as re-writing TOD turns, but their chit-chat is still mostly general responses or based on commonsense knowledge. Kim et al. \shortcite{DBLP:conf/sigdial/KimEGHLH20} propose to insert additional turns into TOD, where the system needs to respond based on the knowledge from domain FAQs. 
The DSTC10 task 2~\cite{DBLP:journals/corr/abs-2109-13489} is based on the dataset from \cite{DBLP:conf/sigdial/KimEGHLH20} with a similar focus. HyKnow~\cite{DBLP:conf/acl/GaoTPLH21} also proposes to insert turns into TOD grounded on knowledge from unstructured documents. These datasets focus on the challenge of detecting those turns requiring external knowledge and selecting the knowledge to generate the responses. In contrast, our dataset focuses on injecting knowledge-grounded chit-chats into the original TOD responses, to make the dialogue more natural and engaging. Our dataset poses more challenges in selecting knowledge based on the dialogue context and generating the responses with both the correct TOD information and the chit-chat seamlessly.

%% file: 03-dataset.tex
\section{The KETOD Dataset}

\subsection{Dataset Construction}
In this section, we describe our framework to construct the KETOD dataset. We start from existing TOD datasets and employ human annotators to augment the functional system responses with knowledge-grounded chit-chat. The proposed approach is demonstrated to give natural, contextual-relevant knowledge enrichment, and meanwhile easy to scale to different datasets. Figure~\ref{fig:data_pipeline} gives an overview of the dataset construction pipeline. 

\noindent \textbf{Data preparation.}
We build upon the SGD dataset~\cite{DBLP:conf/aaai/RastogiZSGK20}, with TOD spanning 16 domains, such as \texttt{Restaurant, Wheather}, etc. Given each TOD, to obtain the knowledge relevant to the dialogue context, we first extract all the entities from the dialogue states and actions. We exclude the domains \texttt{Alarm}, \texttt{Banks}, and \texttt{Payment} as there are mostly no entities involved in these domains; Also, to simplify the human annotation process in the next step, we remove the dialogues with over 10 entities involved.

\noindent \textbf{Knowledge retrieval.}
For each entity, we use the concatenation of the domain name and entity name as the query to retrieve Wikipedia articles. We use the DrQA retriever~\cite{DBLP:conf/acl/ChenFWB17} to retrieve the top 2 Wikipedia articles and take the first 2 paragraphs of each article as the knowledge candidates associated with each entity. Then we break the retrieved articles into sentences, with each sentence as one knowledge snippet. 

\noindent \textbf{Response enrichment.}
In this step, we employ human annotators
to enrich the system responses in the original TOD based on the dialogue context and the retrieved knowledge. 
For each TOD, we present to the annotators the full dialogue, as well as all the knowledge snippets associated with the entities in the dialogue. The annotators can click on each entity name to see the associated knowledge snippets in an expanded textbox. See Appendix A for our annotation interface. 

The annotation process is as follows: 1) Read the full dialog first to have an overall story in mind, as well as the relevant knowledge snippets, then to decide how many turns to enrich with chit-chat and which turn(s) to enrich; If there is no way to make a natural chit-chat enrichment, skip the example. 2) After deciding the turn(s) to enrich with the chit-chat, select the knowledge snippets used to make the enrichment (at most 3 snippets for each turn); 3) Rewrite the system response to enrich with chit-chat grounded on the selected knowledge snippets; The functional information in the original response should be maintained, while may be rephrased to make the enriched response more natural. 

To ensure the dataset quality, we first interview the annotators to select the appropriate hires through a few test examples. Then we launch a training session for all the annotators to learn the task and the annotation interface. We launch the official batches after the annotators can well-master the task. During annotation, we specifically emphasize the contextualization of the knowledge-grounded chit-chat - the enrichment should be contextualized closely on the dialogue context, but not a plain restatement of the knowledge snippets.

\subsection{Dataset Statistics and Analysis}
\label{data_analysis}
We end up with 5,324 dialogues with enriched system responses. We make the split of 4,247/545/532 as the train/dev/test set. Table~\ref{table:gen_stats} shows the statistics of the KETOD dataset. Around 12.1\% of the turns (which indicates mostly 1 or 2 turns in one dialogue) are enriched with knowledge-grounded chit-chat. This intuitively complies with our goal of making the whole dialogue natural and engaging, since too frequent chit-chat may result in redundancy and unnaturalness.  

\noindent \textbf{Quality assessment of the annotation}. During the annotation process, around 12\% of the dialogues cannot be enriched with any turns and thus discarded. It takes around 100 seconds for the annotators to finish each dialogue. To assess the quality of the annotation, we sample 5\% of the annotated dialogues and distribute them to linguistics to check: 1) If the chit-chat enrichment is relevant and natural; 2) If the knowledge snippets are accurately selected corresponding to the enrichment. We end up with a correct rate of 87.0\%. 

\begin{figure}[ht]
\centering
\includegraphics[width=0.48\textwidth]{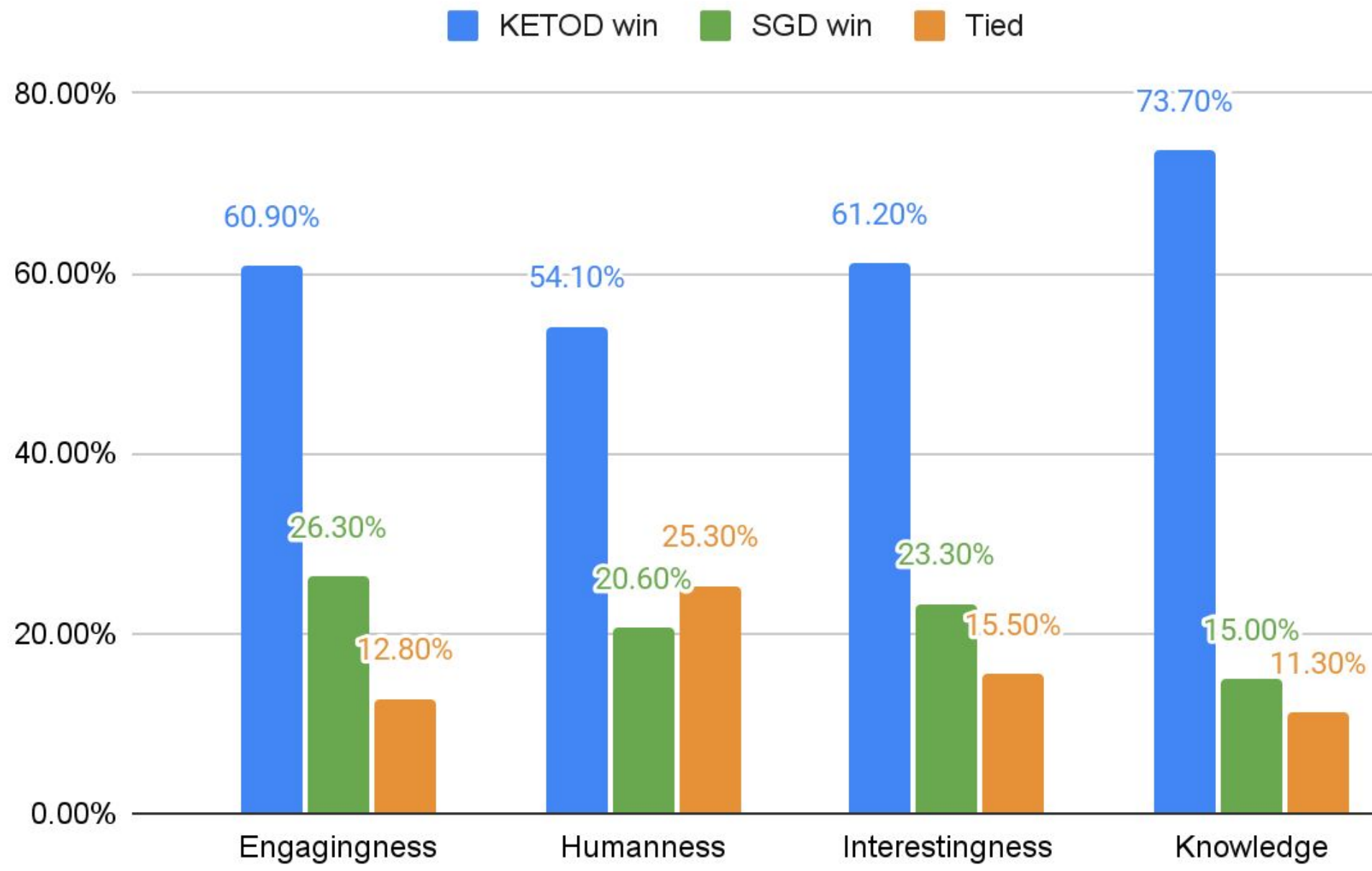}
\caption{Results of pairwise comparison of KETOD vs SGD.} 
\label{fig:human_data}
\end{figure}

\noindent \textbf{Justification of the chit-chat enrichment}. To demonstrate that our proposed knowledge-enriched TOD can be more natural and engaging, we conduct human evaluations to compare KETOD dialogues and their corresponding original TOD dialogues without chit-chat enrichment (SGD). We follow~\cite{DBLP:journals/corr/abs-1909-03087} to make pairwise comparisons of the full dialogues over the following four axes: engagingness, interestingness, knowledge, and humanness. The results in Figure~\ref{fig:human_data} show the superiority of KETOD over all axes. 

\begin{table}[t]
\small
\begin{center}
\resizebox{0.45\textwidth}{!}{%
\begin{tabular}{lr}
\toprule
Dialogues & 5,324\\
Vocabulary & 27k\\
All turns & 52,063\\
Turns enriched with chit-chat & 6,302\\
All entities & 4,639\\
All knowledge snippets & 33,761\\
Avg. \# turns per dialogue & 9.78\\
Avg. \# tokens in enriched responses & 28.07\\
Avg. \# entities per dialogue & 4.98\\
Avg. \# knowledge snippets per dialogue & 70.50\\
\bottomrule
\end{tabular}
}
\caption{General statistics of \textsc{KETOD}.}
\label{table:gen_stats}
\end{center}
\end{table}

%% file: 04-model.tex
\section{Approaches}
In this section, we will describe the proposed two models for the KETOD dataset. 
\begin{figure*}[ht]
\centering
\includegraphics[width=1.0\textwidth]{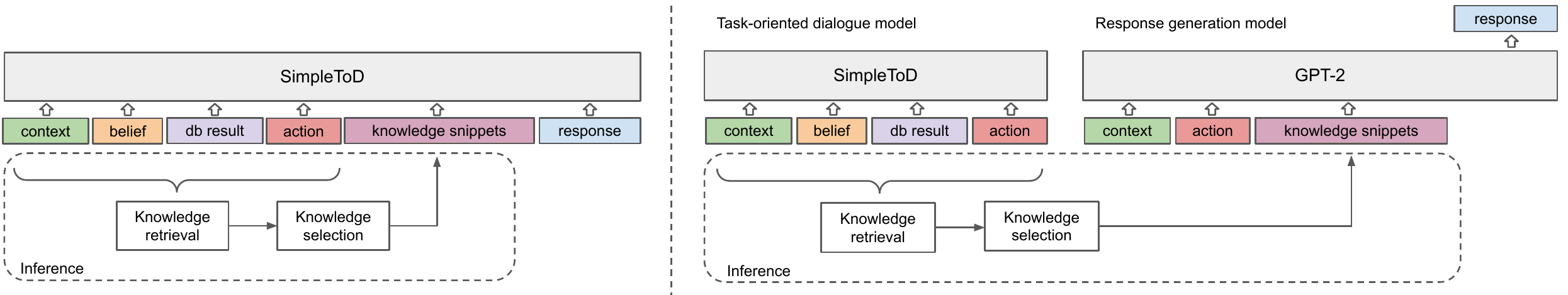}
\caption{Illustration of the models. \textit{Left}: the SimpleToDPlus model; \textit{Right}: the Combiner model; } 
\label{fig:model}
\end{figure*}

\subsection{Overview and Formulations}
For each dialogue turn, denote the dialogue context (history) as $C$, belief states as $B$, database search results as $D$, actions as $A$, the knowledge snippets used for chit-chat enrichment as $K$, the response as $T$. Then we formulate the problem as: given the dialogue context $C$ and a knowledge source (Wikipedia in this dataset), the target is to generate the belief states $B$, actions $A$, and the response $T$, which may be enriched with chit-chat grounded on the knowledge based on the context. The goal of the optimization on KETOD is two-folded: 1) Optimizing the generation of knowledge-enriched responses; 2) Maintaining the task performances; 

In this work, we propose the following modeling framework on KETOD: 1) given the dialogue context, generate the belief states and actions; 2) extract the entities in the belief states and actions, then use these entities to retrieve knowledge candidates (similar as in the dataset construction process); 3) conditioned on the dialogue context, use a knowledge selection model to select knowledge snippets from the knowledge candidates retrieved; 4) generate the knowledge-enriched response conditioned on both the dialogue context and the selected knowledge snippets. 

Based on the above general framework, we propose two architectural approaches, \textbf{SimpleToDPlus} and \textbf{Combiner}, respectively in \S\ref{model1} and \S\ref{model2}. 

\subsection{Knowledge Selection}
\label{kg_select}
After the generation of belief states and actions, we retrieve the knowledge snippet candidates from Wikipedia using the entities in the belief states and actions. The average number of knowledge snippets candidates retrieved for each dialogue is around 70. 
It is impractical to input all of them into the models due to the large amount. As we have the annotation for the ground truth knowledge snippets used for each chit-chat enrichment, we train a knowledge selection model to select the top knowledge snippets most appropriate for chit-chat enrichment. Specifically, we concatenate the dialogue context with each knowledge snippet as the input. Then we use BERT~\cite{DBLP:conf/naacl/DevlinCLT19} to train a simple classifier to rank all the knowledge snippets candidates. We take the top 3 ones as the knowledge selection results. We use the same knowledge selection model for both architectures.

\subsection{SimpleToDPlus}
\label{model1}
SimpleToD~\cite{DBLP:conf/nips/Hosseini-AslMWY20} is a recent popular approach on TOD, which uses one single language model to sequentially generate the belief states, actions, and responses. It has achieved strong performances in all the above functional tasks. In this work, we propose its extension, \textbf{SimpleToDPlus}, to generate knowledge-enriched responses for TOD. The left part of Figure~\ref{fig:model} shows the overview of SimpleToDPlus. 
We formulate the training sequence as:
\begin{equation}
    \text{[} C, B, D, A, K, \text{<chitchat>}, T \text{]}
\end{equation}
Where <chitchat> is a tag to indicate the decision of whether to enrich the response with knowledge grounded chit-chat or not. If the response is not enriched, we insert the tag <nochitchat>. Since the number of the gold knowledge snippets varies from 1 to 3 (as in the dataset construction), to be compatible with inference time, here we first run the knowledge selection model on all training instances. Then we construct the knowledge snippets $K$ as the merge of the gold knowledge snippets and the knowledge selection model results, truncated to 3 ones. If the response is not enriched with chit-chat, i.e., no gold knowledge snippets, we still put 3 snippets from the knowledge selection model ranking results here during training. 
 
In the inference time, we first sequentially generate the belief states and actions. Then we extract the entities from the generated belief states and actions, and apply the same process of knowledge retrieval as in dataset construction. Next, we run the knowledge selection model on the retrieved knowledge candidates and take the top 3 knowledge snippets as the model input followed by the generated actions. At last, the model generates the decision to make chit-chat enrichment or not, followed by the final response.
 
Since the knowledge-enriched response is conditioned on the entity knowledge from the belief states and actions, we need to directly include the entities in the actions and responses during generation, instead of generating a delexicalized result first and then lexicalizing in the post-process as in the original SimpleToD. To simplify, we use the oracle database search results for all the experiments. 

\begin{table*}[t]
\small
\begin{center}
\resizebox{0.8\textwidth}{!}{%
\begin{tabular}{lcccccc}
\toprule
\textbf{Models} & \textbf{Joint GA} & \textbf{Avg GA} & \textbf{Act-Slot F1} & \textbf{BLEU-4\textsubscript{aug}} & \textbf{BLEU-4\textsubscript{orig}} & \textbf{BLEU-4\textsubscript{all}} \\
\midrule
SimpleToD-ref & 27.6 & 54.2 & 67.6 & - & - & - \\
\midrule
SimpleToD & 23.7 & 50.1 & 62.7 & 4.8 & 10.7 & 10.0 \\
SimpleToDPlus & \textbf{28.6} & \textbf{52.2} & \textbf{66.9} & 6.3 & \textbf{11.7} & \textbf{11.0} \\
Combiner & 24.5 & 51.5 & 64.5 & \textbf{6.5} & 9.9 & 9.5 \\
\bottomrule
\end{tabular}
}
\caption{Main experiment results: Both SimpleToDPlus and Combiner outperform the baseline. Overall SimpleToDPlus obtains better response generation performance while maintaining competitive TOD performance. }
\label{table:main_res}
\end{center}
\end{table*}

\subsection{Combiner}
\label{model2}
SimpleToDPlus models all the generations in an end-to-end manner. In \textbf{Combiner}, we use a pipeline of a TOD model followed by a response generation model to separate the TOD part (belief states, actions) with the generation of knowledge-enriched responses. The goal is to study whether an independent model can better learn each task with less interference from the other. The overview of the architecture is shown on the right of Figure~\ref{fig:model}. 

For the TOD model, we use SimpleToD to generate the belief states and actions, with the training sequence as:
\begin{equation}
    \text{[} C, B, D, A\text{]}
\end{equation}
We find that including the knowledge-enriched responses during training degrades the task performance, indicating the disturbance from the ungrounded knowledge in the responses. 

For the response generation model, we use GPT-2~\cite{radford2019language} with the concatenation of the dialogue context, actions, and the knowledge snippets as the prompt:
\begin{equation}
    T = \text{GPT-2}(C, A, K)
\end{equation}
We use the same way of constructing the merged knowledge snippets during training, and the same process of knowledge retrieval and selection during inference as in SimpleToDPlus. 


%% file: 05-experiments.tex
\section{Experimental Results}

\noindent \textbf{Baseline model}.
We use SimpleToD~\cite{DBLP:conf/nips/Hosseini-AslMWY20} as our baseline model, i.e., with the training sequence as \text{[} $C, B, D, A, T$ \text{]}, without the injection of knowledge snippets. Therefore the knowledge-grounded chit-chat in the responses $T$ do not have any knowledge groundings - we aim to show the necessity of knowledge grounding for our task, as well as the effectiveness of our proposed models to incorporate knowledge.  

\noindent \textbf{Experimental setups and evaluations.}
Check Appendix B for details of model training and parameter settings.
For the TOD performances, we evaluate the belief states with joint goal accuracy (Joint GA) and average goal accuracy (Avg GA), and the actions with act-slot F1, same as \cite{DBLP:conf/naacl/SunMCRSLWLCC21}. For the automatic evaluations of response generation, we use three BLEU-4 scores: BLEU-4\textsubscript{aug} for evaluating the responses enriched with knowledge; BLEU-4\textsubscript{orig} for evaluating the responses not enriched with knowledge; BLEU-4\textsubscript{all} for evaluating all responses;

\subsection{Main Results}
\noindent \textbf{Performance on response generation}. Table~\ref{table:main_res} shows our main experiment results. For the performances on response generation, we can see that both of our proposed models, SimpleToDPlus and Combiner, improve on the knowledge-enriched response generation (BLEU-4\textsubscript{aug}) over the SimpleToD baseline. Since in the baseline, we do not include the knowledge snippets in the input, the generated responses are mostly enriched with random knowledge or frequent knowledge in the training data. The improvements demonstrate the necessity of knowledge grounding and the effectiveness of the proposed knowledge enrichment methods. Combiner performs slightly better on knowledge-enriched responses than SimpleToDPlus but falls short on the responses without knowledge-enrichment (i.e., original TOD responses). This is partially because of its pipeline nature - a separated response generation module can better learn the knowledge enrichment without the disturbance of other tasks, but the error cascading from the generated actions degrades the performance of the TOD responses part. 

\noindent \textbf{Performances on belief states and actions}. To better study how the knowledge enrichment affects the TOD performances, we first train SimpleToD on our dataset without the knowledge enrichment, i.e., replace all the knowledge-enriched responses with the original responses in SGD. We name it as SimpleToD-ref in Table~\ref{table:main_res}, serving as a reference of the original TOD performances. The SimpleToD baseline gives largely degraded performances due to the disturbance from the ungrounded knowledge in the responses during training. Therefore in Combiner, we do not include the responses in the training sequences of the TOD model (specified in section~\ref{model2}), and obtain better scores. SimpleToDPlus achieves the best TOD performances, which are nearly competitive with SimpleToD-ref. This is partially due to the enhancement of language modeling ability brought by the training on the responses grounded on the input knowledge.

\noindent \textbf{Human evaluations}.
In order to get the more comprehensive measure of the response generation performances, we conduct human evaluations for both dialogue-level pairwise comparison and turn-level factualness evaluation. For dialogue-level pairwise comparison, we randomly sample 200 dialogues from the test set and apply the same process as in dataset evaluation (\ref{data_analysis}). For each model, we construct the full dialogue results by concatenating the generated response for each turn given the gold dialogue context. Table~\ref{table:human_res_model} shows the results of pairwise comparison between the SimpleToDPlus model and the Combiner model, demonstrating SimpleToDPlus is more performant. Table~\ref{table:human_res_gold} shows the results of pairwise comparison between SimpleToDPlus and the gold reference, indicating there is still a large room for further improvements. See Appendix C for the human evaluation results of comparing both methods to the baseline. For turn-level factualness evaluation, we randomly sample one turn with chit-chat enrichment from each dialogue, and present both the generated response and the selected knowledge snippets to the annotators. The annotators are asked to check whether the chit-chat in the responses are factually correct based on the knowledge snippets. SimpleToDPlus and Combiner obtain the factualness correct rate of 64.2\% and 66.1\%, respectively. In summary, Combiner achieves better factualness of knowledge enrichment since its independent response generation model can better focus on the learning of knowledge groundings. But its error cascading due to the pipeline nature may degrade the overall consistency and human-likeness of the generated dialogue.

As we have two optimization goals on KETOD 1) Optimizing the generation of knowledge-enriched responses; 2) Maintaining the task performances, we consider SimpleToDPlus as a better model regarding the overall performances. We will use the results of SimpleToDPlus for the ablations and other analyses in the rest of the  experiments. 
\begin{table}[t]
\small
\begin{center}
\resizebox{0.45\textwidth}{!}{%
\begin{tabular}{lccc}
\toprule
\textbf{Metrics} & \textbf{\makecell{SimpleToDPlus win \\ (\%)}} & \textbf{\makecell{Combiner win \\ (\%)}} & \textbf{\makecell{Tied \\ (\%)}} \\
\midrule
Engagingness & 47.8 & 24.5 & 27.8 \\
Interestingness & 34.5 & 19.0 & 46.5 \\
Knowledge & 29.5 & 26.3 & 44.3 \\
Humanness & 43.3 & 23.8 & 33.0  \\
\bottomrule
\end{tabular}
}
\caption{Human evaluation of SimpleToDPlus vs. Combiner. }
\label{table:human_res_model}
\end{center}
\end{table}
\begin{table}[t]
\small
\begin{center}
\resizebox{0.45\textwidth}{!}{%
\begin{tabular}{lccc}
\toprule
\textbf{Metrics} & \textbf{\makecell{SimpleToDPlus win \\ (\%)}} & \textbf{\makecell{Gold win \\ (\%)}} & \textbf{\makecell{Tied \\ (\%)}} \\
\midrule
Engagingness & 16.8 & 60.5 & 22.8 \\
Interestingness & 12.0 & 51.0 & 37.0 \\
Knowledge & 14.5 & 44.8 & 40.8 \\
Humanness & 17.3 & 58.0 & 24.8  \\
\bottomrule
\end{tabular}
}
\caption{Human evaluation of SimpleToDPlus vs. Gold. }
\label{table:human_res_gold}
\end{center}
\end{table}
\begin{table}[t]
\small
\begin{center}
\resizebox{0.45\textwidth}{!}{%
\begin{tabular}{lcc}
\toprule
\textbf{} & \textbf{BLEU-4\textsubscript{aug}} & \textbf{BLEU-4\textsubscript{all}} \\
\midrule
\multicolumn{2}{l}{\textbf{Given gold TOD results, decision, and knowledge}} & \\
\midrule
SimpleToD & 6.5 & 13.1 \\
SimpleToDPlus & 9.7 & 14.6 \\
Combiner & 14.6 & 15.1 \\
\midrule
\multicolumn{2}{l}{\textbf{Given gold TOD results}} & \\
\midrule
SimpleToD & 6.3 & 12.8 \\
SimpleToDPlus & 7.4 & 14.0 \\
Combiner & 9.6 & 13.9 \\
\bottomrule
\end{tabular}
}
\caption{Analysis of different inference stages: we provide the models with gold results up to certain stages, and investigate the performances for the inferences on following stages.}
\label{table:res_stage}
\end{center}
\end{table}

\begin{table}[t]
\small
\begin{center}
\resizebox{0.45\textwidth}{!}{%
\begin{tabular}{lccc}
\toprule
\textbf{} & \textbf{BLEU-4\textsubscript{aug}} & \textbf{BLEU-4\textsubscript{all}} & \textbf{\makecell{Knowledge selection \\ recall (\%)}} \\
\midrule
Gold & 9.7 & 14.6 & 100.0 \\
BERT selection & 7.8 & 14.4 & 52.7 \\
TF-IDF selection & 6.6 & 13.7 & 14.1 \\
\bottomrule
\end{tabular}
}
\caption{SimpleToDPlus response generation performance with varying knowledge selection strategies.}
\label{table:res_kg}
\end{center}
\end{table}
\begin{table}[t]
\small
\begin{center}
\resizebox{0.45\textwidth}{!}{%
\begin{tabular}{lccc}
\toprule
\textbf{} & \textbf{BLEU-4\textsubscript{aug}} & \textbf{BLEU-4\textsubscript{all}} & \textbf{\makecell{Enrichment decision \\ F1 (\%)}} \\
\midrule
Gold decision& 9.7 & 14.6 & 100.0 \\
Predicted decision & 8.0 & 14.1 & 58.7 \\
\bottomrule
\end{tabular}
}
\caption{SimpleToDPlus response generation performance using (1) the gold set of turns to enrich with chit-chat, and (2) the predicted set of turns.}
\label{table:res_decision}
\end{center}
\end{table}
\begin{table}[t]
\small
\begin{center}
\resizebox{0.45\textwidth}{!}{%
\begin{tabular}{lccccc}
\toprule
\textbf{} & \textbf{All} & \textbf{Hotels} & \textbf{Movies} & \textbf{Restaurant} & \textbf{Music}\\
\midrule
BLEU-4\textsubscript{aug} & 6.3 & 7.1 & 5.2 & 5.1 & 7.7 \\
BLEU-4\textsubscript{all} & 11.0 & 10.3 & 12.2 & 14.0 & 12.3 \\
\bottomrule
\end{tabular}
}
\caption{Domain breakdown of SimpleToDPlus response generation performances.}
\label{table:res_domain}
\end{center}
\end{table}

\begin{figure*}[t]
\centering
\includegraphics[width=0.96\textwidth]{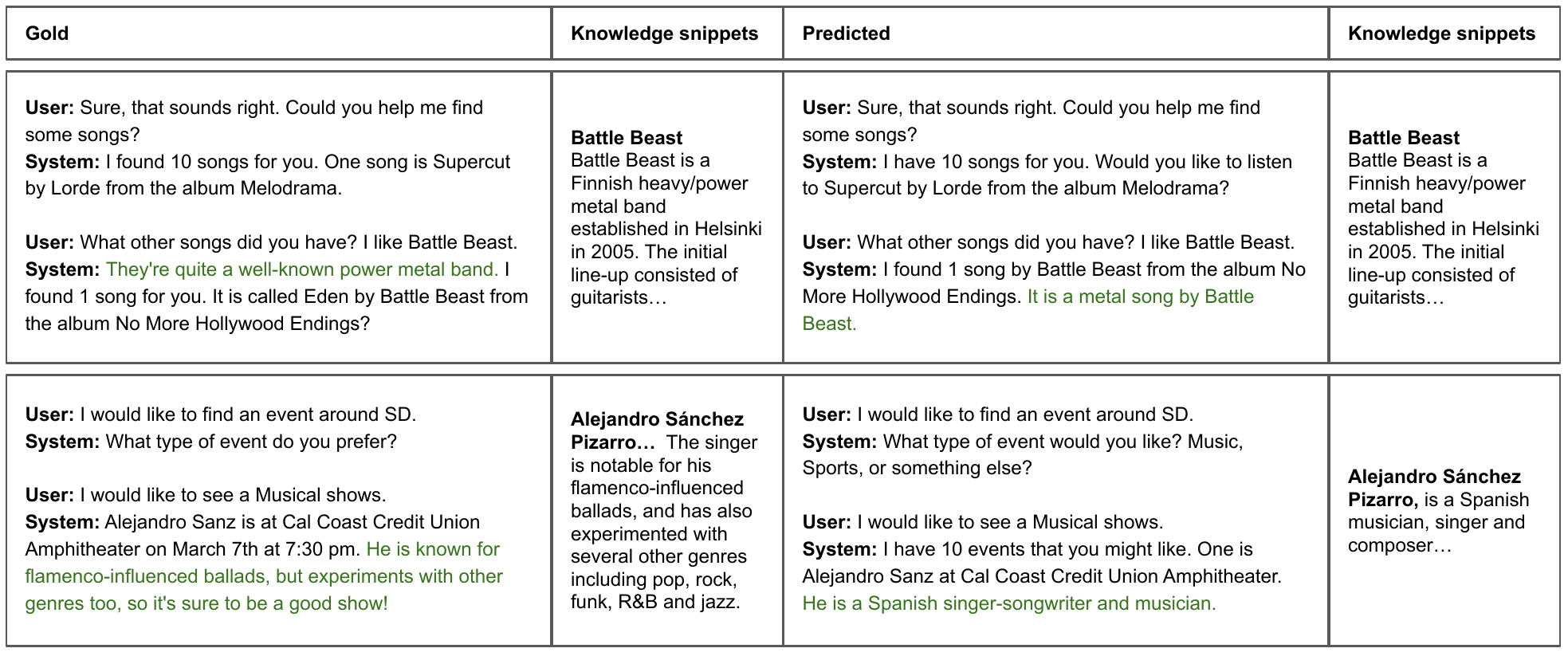}
\caption{Case studies: two examples of comparing the generation from SimpleToDPlus (right) with the gold reference (left), together with the knowledge snippets selected. Overall our model can mostly generate reasonable knowledge enrichment, but still falls short on engagingness and consistency compared to the gold references.} 
\label{fig:case}
\end{figure*}
\subsection{Ablations and Analysis}
\noindent \textbf{Analysis of different inference stages.}
There are several inference stages for this task - the TOD results (belief states and actions), the selection of knowledge snippets, and the final response generation, where each stage is conditioned on previous results. Therefore the errors accumulate through all the stages leading to the final performances. Here we run another two sets of experiments to study such error accumulations and compare the two models. Specifically, first, we feed the models with the gold TOD results, chit-chat decisions, and knowledge snippets, to solely test the abilities to generate the knowledge-enriched responses; Second, we feed the models with the gold TOD results to test the following stages of knowledge selection and the response generation. The results are shown in Table~\ref{table:res_stage}. Compared with the full inference results in Table~\ref{table:main_res}, we can see that the Combiner model largely outperforms SimpleToDPlus if provided with more gold results for previous stages. However, it gradually falls behind SimpleToDPlus when moving to fully end-to-end inference due to the error cascading of its pipeline nature. 

\noindent \textbf{Importance of knowledge selection strategies.}
To demonstrate the importance of the knowledge selection strategies (and their subsequent recall performance), we run SimpleToDPlus with 1) gold knowledge snippets; 2) predicted knowledge snippets (with BERT); 3) knowledge snippets selected by heuristics (we use TF-IDF matching between the current dialogue turn and the knowledge snippets). To eliminate the influences brought by other inference stages, we feed the model with gold TOD results (dialogue states and actions). The results are shown in Table~\ref{table:res_kg}. There exists a certain level of variance for knowledge selection, e.g., when recommending a song for the user, you may talk about its genre, its singer, or the album.  

\noindent \textbf{Learning \textit{when} to inject knowledge-enriched chit-chat.} 
In all models, we use the special token `<chitchat>' and `<nochitchat>' to indicate the decision to inject knowledge enrichment for the responses. To study the effect of the chit-chat injection decision-making accuracy on the overall dialogue tasks, we run SimpleToDPlus (1) with the ground-truth information of turns to enrich with chit-chat, and (2) with the predicted decisions, using the gold TOD results. Table~\ref{table:res_decision} shows the performance gap, which highlights the importance of knowing \textit{when} to inject knowledge-enriched chit-chat. While such decisions are conditioned on the dialogue history, e.g., we may tend to not enrich a turn if many of the previous turns are enriched to avoid redundancy, there also exists some variance. In a real system, we may consider specifying the turns to make the chit-chat enrichment instead of letting the model make the decision. 

\noindent \textbf{Domain analysis.}
We investigate the model performance for each domain in Table~\ref{table:res_domain}. We observe that the performance differences may depend on the variance of the enriched knowledge.
Domains with larger variance on selected knowledge tend to have lower automatic scores. For example, in \texttt{Hotels} domain, mostly the chit-chat is about the locations since there are mostly location entities involved in this domain. But for the \texttt{restaurants} domain, the enriched knowledge can be about the food, the restaurant, as well as the location. The selected knowledge shows more diversity and variance. 

We provide case studies in Figure~\ref{fig:case} to compare the predicted results with the gold references. 

%% file: 06-conclusion.tex
\section{Conclusion}
In this work, we propose to combine task-oriented dialogue with knowledge-grounded chit-chat, and construct a new dataset named KETOD, with manually composed knowledge-enriched system responses. We conduct comprehensive experiments on our new dataset to study the insights and challenges. We believe that our proposed task is an important step towards the ultimate goal to build a unified, human-like conversational AI. Our new dataset KETOD, annotated by experts, will greatly facilitate the research in this direction.


%% file: 07-ethics.tex
\section{Ethical Considerations}
\noindent \textbf{Data Access and Licensing.} We develop the KETOD dataset based on the publicly available SGD dataset\footnote{https://github.com/google-research-datasets/dstc8-schema-guided-dialogue}~\cite{DBLP:conf/aaai/RastogiZSGK20}. The SGD dataset is publicly available under the CC-BY-SA-4.0 License. 

\noindent \textbf{Dataset Collection Process and Conditions.}
This project is approved by our Institutional Review Board (IRB). Our annotators are all U.S. based. For the annotation of our KETOD dataset, linguistics for assessing data quality, and all the human evaluations, our annotators were hired
as full-time employees through a leading annotation services vendor, and were paid in accordance with a fair wage rate. During the data annotation, we instruct the annotators to skip any example that contains offensive or any unethical contents. 


%% file: 08-appendix.tex
\section*{Appendix A: Dataset Construction}
\begin{figure*}[ht]
\centering
\includegraphics[width=0.96\textwidth]{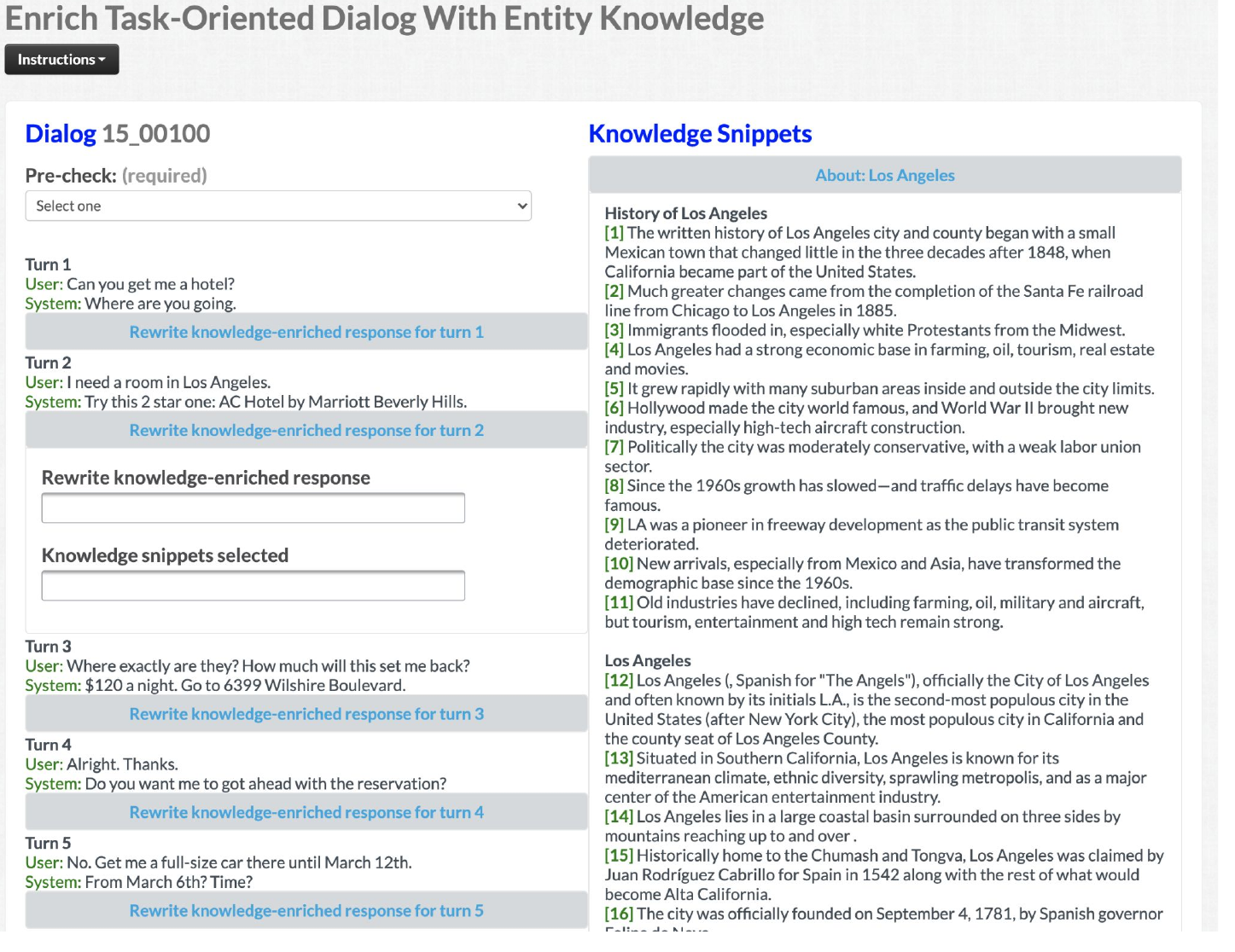}
\caption{Our annotation interface example 1. } 
\label{fig:ann_1}
\end{figure*}

\begin{figure*}[ht]
\centering
\includegraphics[width=0.96\textwidth]{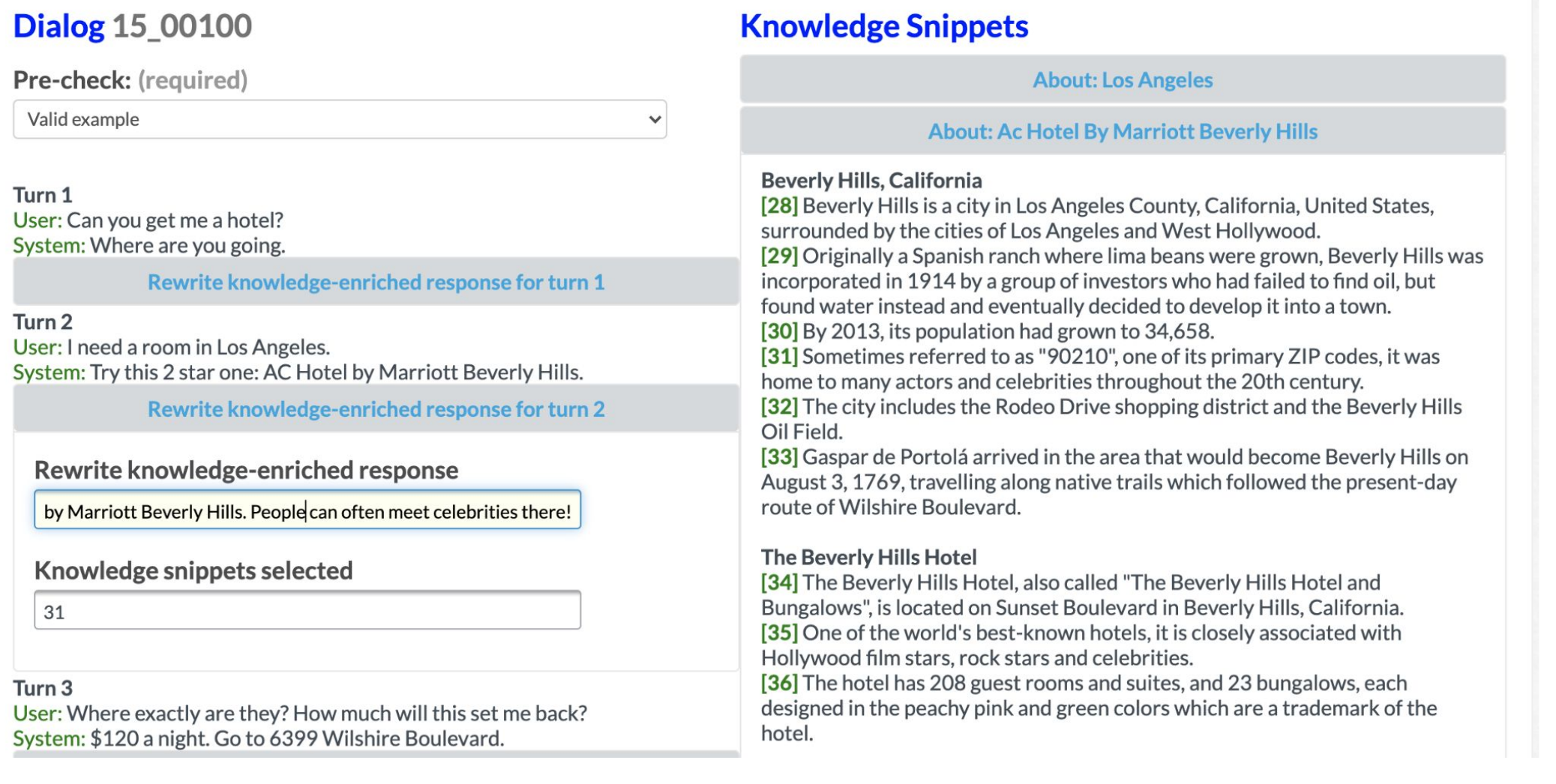}
\caption{Our annotation interface example 2. } 
\label{fig:ann_2}
\end{figure*}

Figure~\ref{fig:ann_1} shows our annotation interface to add knowledge-grounded chit-chat to TOD. The left part shows the full dialogue, where the annotators can click and expand each turn to make the chit-chat enrichment. The right part shows all the entities with the associated knowledge snippets. The annotators can click on each entity name to expand the textbox to see the knowledge snippets. We add index number to each knowledge snippet (shown in green brackets), and the annotators are asked to write down the indexes of the knowledge snippets they used for writing the knowledge grounded chit-chat. Figure~\ref{fig:ann_2} shows one example annotation turn using our interface.


\section*{Appendix B: Model and Training Details}
All the implementations are based on the Huggingface Transformers library\footnote{https://github.com/huggingface/transformers}. For all models, we use the Adam optimizer~\cite{DBLP:journals/corr/KingmaB14}.
For the knowledge selection model, we use BERT-base with learning rate of 3e-5 and batch size of 16. For the baseline SimpleToD model, SimpleToDPlus model, and Combiner model, we all use learning rate of 1e-4 and batch size of 16. All the experiments are done using TESLA M40 GPU cards. 

\section*{Appendix C: Evaluation Details}
Table~\ref{table:human_res_1base} and~\ref{table:human_res_2base} show the human evaluation results of SimpleToDPlus vs. SimpleToD, and Combiner vs. SimpleToD, respectively. 
\begin{table}[t]
\small
\begin{center}
\resizebox{0.45\textwidth}{!}{%
\begin{tabular}{lccc}
\toprule
\textbf{Metrics} & \textbf{\makecell{SimpleToDPlus win \\ (\%)}} & \textbf{\makecell{SimpleToD win \\ (\%)}} & \textbf{\makecell{Tied \\ (\%)}} \\
\midrule
Engagingness & 40.0 & 30.3 & 29.8 \\
Interestingness & 31.8 & 19.5 & 48.8 \\
Knowledge & 38.0 & 18.3 & 43.8 \\
Humanness & 38.3 & 26.8 & 35.0  \\
\bottomrule
\end{tabular}
}
\caption{Human evaluation of SimpleToDPlus vs. SimpleToD. }
\label{table:human_res_1base}
\end{center}
\end{table}

\begin{table}[t]
\small
\begin{center}
\resizebox{0.45\textwidth}{!}{%
\begin{tabular}{lccc}
\toprule
\textbf{Metrics} & \textbf{\makecell{Combiner win \\ (\%)}} & \textbf{\makecell{SimpleToD win \\ (\%)}} & \textbf{\makecell{Tied \\ (\%)}} \\
\midrule
Engagingness & 34.8 & 33.5 & 31.8 \\
Interestingness & 27.0 & 22.5 & 50.5 \\
Knowledge & 32.5 & 23.0 & 44.5 \\
Humanness & 27.8 & 32.5 & 39.8  \\
\bottomrule
\end{tabular}
}
\caption{Human evaluation of Combiner vs. SimpleToD. }
\label{table:human_res_2base}
\end{center}
\end{table}